\documentclass[conference]{IEEEtran}
\IEEEoverridecommandlockouts
\usepackage{cite}
\usepackage{amsmath,amssymb,amsfonts}
\usepackage{algorithmic}
\usepackage{graphicx}
\usepackage{textcomp}
\usepackage{xcolor}

\def\BibTeX{{\rm B\kern-.05em{\sc i\kern-.025em b}\kern-.08em
    T\kern-.1667em\lower.7ex\hbox{E}\kern-.125emX}}
\begin{document}

\title{ \textbf{\textit{Minimal-Configuration Anomaly Detection for IIoT Sensors}} \huge \\ 
{ \large \textbf{\textit{A Systematic Analysis of Deep Learning Approaches}} \vspace{2em}}}

\author{\IEEEauthorblockN{Clemens Heistracher}
\IEEEauthorblockA{Center for Digital Safety and Security \\
Austrian Institute of Technology \\
}
\and
\IEEEauthorblockN{Anahid Jalali}
\IEEEauthorblockA{Center for Digital Safety and Security \\
Austrian Institute of Technology \\
}
\and
\IEEEauthorblockN{Axel Suendermann}
\IEEEauthorblockA{Distributed AI Systems Research Group \\
Siemens Technology Austria \\
}\and
\and
\IEEEauthorblockN{Sebastian Meixner
}
\IEEEauthorblockA{Distributed AI Systems Research Group \\
Siemens Technology Austria \\
}\and
\IEEEauthorblockN{Daniel Schall}
\IEEEauthorblockA{Distributed AI Systems Research Group \\
Siemens Technology Austria \\
}\and
\IEEEauthorblockN{Bernhard Haslhofer}
\IEEEauthorblockA{Center for Digital Safety and Security \\
Austrian Institute of Technology \\
}
\and
\IEEEauthorblockN{Jana Kemnitz}
\IEEEauthorblockA{Distributed AI Systems Research Group \\
Siemens Technology Austria \\
}

}


\maketitle

\begin{abstract}

The increasing deployment of low-cost IoT sensor platforms in industry boosts the demand for anomaly detection solutions that fulfill two key requirements: minimal configuration effort and easy transferability across equipment.
Recent advances in deep learning, especially long-short-term memory (LSTM) and autoencoders, offer promising methods for detecting anomalies in sensor data recordings.
We compared autoencoders with various architectures such as deep neural networks (DNN), LSTMs and convolutional neural networks (CNN) using a simple benchmark dataset, which we generated by operating a peristaltic pump under various operating conditions and inducing anomalies manually.
Our preliminary results indicate that a single model can detect anomalies under various operating conditions on a four-dimensional data set without any specific feature engineering for each operating condition. We consider this work as being the first step towards a generic anomaly detection method, which is applicable for a wide range of industrial equipment.
\end{abstract}


\section{Introduction}\label{intro}

Prognostics and health management approaches have been studied extensively across industrial applications, such as aircraft engines\cite{Tolani2006}, wind turbines\cite{Zaher2009}, and other expensive and mission critical machines. The application of IIoT sensors \cite{Tong2019, Zhao2017, Bierweiler2019} enables continuous monitoring on previously unequipped industrial assets. The SITRANS multi sensor \cite{Bierweiler2019} enables affordable to monitor equipment and industry expects huge cost savings from the implementation of data-driven predictive maintenance techniques such as anomaly detection.
However, the implementation of a traditional anomaly detection technique for some specific equipment often requires significant manual feature engineering \cite{Su2019, Wang2020} and model optimization effort. Alternative approaches requiring less effort and transferability to similar equipment are therefore becoming increasingly relevant in industrial contexts. A number of anomaly detection methods for predictive maintenance have been proposed.  Kato et al. \cite{kato2001integrating} proposed a rule-based approach for fault detection in spacecrafts. Principal component analysis (PCA) based anomaly detection in networks was discussed by \cite{Camacho2016}. \cite{Sakurada2014} showed that autoencoders can outperform PCA based approaches for telemetry data of spacecrafts. Unsupervised anomaly detection with deep autoencoders was shown by \cite{zong2018deep, MALEKI2021}.

Our aim is to develop an unsupervised anomaly detection method for a universally deployable IIoT sensor tag, which records multivariate data. It should learn anomalies automatically over time and thereby reduce manual feature engineering effort. Our specific contributions so far are: 
(i) define initial requirements and derive design rationals for minimal-configuration anomaly detection for IIoT Sensors (ii) provide a hand-crafted benchmark data set of evaluating anomaly detection approaches, and 
(iii) train various deep neural networks with autoencoder architectures and evaluated them against benchmark models.
\section{Industrial Requirements and Design Rational}\label{design}

A low-cost multi sensor should be applicable to any industrial asset. With the help of this multi- sensor, the conditions should be monitored without any meta information available. For this purpose, the healthy state with all typical operational conditions is recorded with and used as a reference for anomaly detection. The amount of healthy training data is not strictly limited and several days can be expected. The system should be minimal configurable. The only input parameter is the healthy reference data. The system should be operable by a non-machine learning expert. The user should be a domain expert and select a time period with typical operational conditions as reference. Everything else is left to the model and the system. The decision for an unsupervised machine learning paradigm results from the requirements. 

\section{Data Set Creation}\label{dataset}

We have selected the peristaltic pump as it can be operated using various operating conditions and anomalies in the pipe's water flow resistance can be applied easily. Further, the rotor of a peristaltic pump is representative for many rotary equipment in the industry, such as fans, compressors and turbines. Additionally, degradation is commonly observed and the can be controlled as pipes can be replaced easily. Abrasion of the pipes will be used to predict failure of the pipe system in subsequent work. Peristaltic pumps are used for sterile or aggressive liquids, as the pump doesn't contact the fluid. The flow of liquids is induced by a repeating sequential compression of a flexible tube that pushed the liquid in one direction.


\begin{figure}[h]
\centering
\includegraphics[width=.45\textwidth]{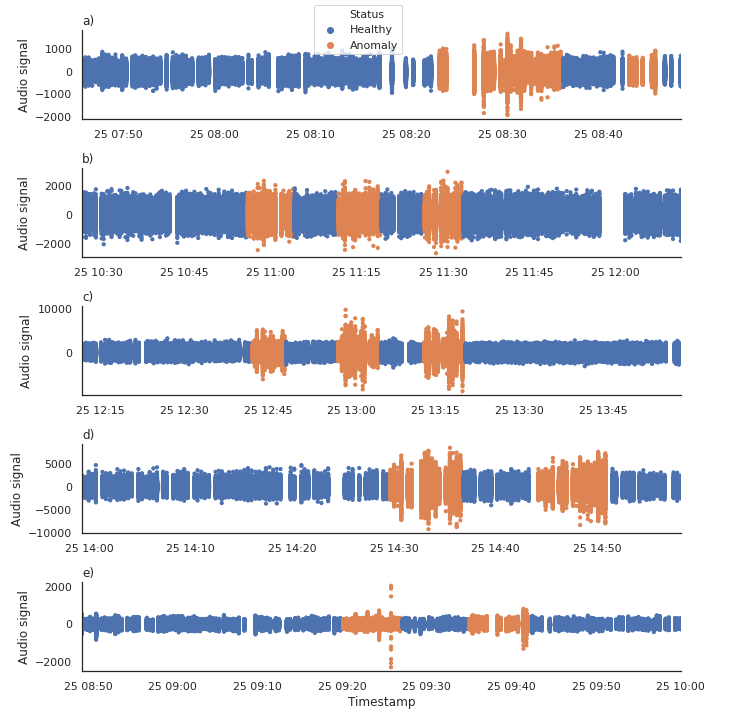}
\caption{The audio signal for various operating conditions: a) 100  Hz, b) 150   Hz, c) 200   Hz, d) 250  Hz, e) 50  Hz with an additional 12 hours of normal state. }
\label{fig:dataset}
\end{figure}

We created our data set with a prototype of the SIEMENS SITRANS multi sensor specifically developed for industrial applications and harsh environments \cite{Bierweiler2019}. The sensor offers a multiple number of measurement parameters. In this work, we used the sensor tag that records three-axis vibration, each with a frequency of $6664 \  Hz$ , an audio signal with $16k \ Hz$, and temperature. Due to restrictions in bandwidth, vibration and audio are measured sequentially for 1024 data points every $60 \ s$. We mounted the sensor tag on at rotational axis of the pump's rotor and documented the sensor's angle and its horizontal axis. To simulate various operating conditions, we operated the pump under various conditions by changing the pump's frequency. Further, we induced anomalies by restricting the water flow in the tube leading to the pipe. We scheduled the data acquisition to generate a data set that is balanced for operating conditions and anomalies. Additionally, we documented the replacement of the tube to allow analysis of the tube's degradation and we also perform measurements with a rotated sensor to evaluate the models robustness against rotation. A model, which performs well if the sensor was rotated and reattached, is a candidate for architectures that are easily transferable across equipment and require minimal configuration effort.  The data set contains 3041 samples with each 1024 data points for audio and the three vibration axis and will be made publicly available.

\section{Experimental Setup}\label{experiment}

We trained unsupervised machine learning models to detect the anomalies in our dataset by using autoencoders based on a fully connected deep neural network (DNN), long short-term memory (LSTM) networks and convolutional neural networks (CNN). Autoencoder networks are trained to reproduce a input signal by minimization the error between input and output signal, which is called the reconstruction error. This is done by setting the input values as the target values. If there is a layer with a feature space lower than the input space, the autoencoder is forced to learn a compressed representation and therefore needs to generalize and approximate the input. In other words, a bottleneck in the network requires the encoder to extract the most substantial information.

For anomaly detection, autoencoders are trained to reconstruct only healthy machine data. It is assumed that the autoencoder learns to reconstruct the input for healthy machine data, as it was trained to do so, but will fail to reconstruct anomaly data. The reconstruction error --- the error between input and output signal --- can be used as an anomaly score. A reconstruction error above a threshold indicates an anomaly. The threshold is calculated on a subset of the healthy data that was excluded from training, by calculation mean + standard deviation of the reconstruction error on the subset. We evaluated the effectiveness of the anomaly detection using standard accuracy (Ac.), precision (P), recall (R) and the F1-score. 

We compared the performance of our models on a variety of features. We used the audio and the vibration (vib. 3D) signal separately as well as a combinations of both. Further, we use raw signal and the fast Fourier transform (FFT) of the signals. 
In order to achieve invariance towards rotation of the sensor, we also use the euclidean norm of the three-dimensional vibration signal, which is denoted as vib. 1D. Feature names are generated from the options mentioned in this paragraph and are shown in Table \ref{table_results}. For example, "vib. 1D $\&$ audio" denotes the raw audio signal in combination with the euclidean norm of the raw vibration signal. Feature vectors are then min-max normalized based on values of the train set. 

The DNN model consists of six fully connected layers of varying dimension. With $x$ being the length of the input signal and $n$ being the characteristic number of neurons, the layers are of dimensions $x, \ n, \  \frac{n}{3}, \  \frac{n}{4}, \  \frac{n}{3}, \ n, \ x $. All units use the tanh activation function and $n$ is selected form the range $64$ to $200$, depending on the selected feature. For multivariant features and DNN we stack the feature vectors to achieve one-dimensional features, whereas CNN and LSTM operate directly on the two-dimensional features. 
Due to the high feature size (1024) compared to the number of samples, we use a rolling window of dimension 64 to create more and smaller feature vectors, that make a smaller model possible. The models threats each sub-vector independently and a decision for a original sample is created by majority vote on the sub-vectors from the rolling window.

The LSTM models consists of stacked LSTMs networks, with increasing dimension that create a two dimensional output by returning an output for every time step. The bottleneck layer reduces the dimension by returning only the last output and is followed by a repetition of the last output for every time step. Then the number of neurons is decreased opposed to the encoder. With $n$ being the number of units, the dimensions of the layers are  $n, \ \frac{n}{2},\ \frac{n}{4}, \ \frac{n}{16}, \ \frac{n}{16}, \ \frac{n}{4}. \ \frac{n}{2}, \ n$ (rounded) with a lower cap of $16$ and $n = 150$.

The CNN's encoder consists of an alternating sequence of convolutional and max pooling layers, each of dimension two, with the number of filters for the layers in the encoder being $16, \ 32, \ 64, \ 128$. A fully connected layer as a bottleneck and a reversed encoder as a decoder. 

We have implemented an end to end pipeline to evaluate the key requirement: ``minimal configuration effort''.
Further, we compare our model to simple statistical benchmarks based on the reconstruction error using principal component analysis (BM PCA) (see. ~\cite{Camacho2016}) and an approach similar to boxplots, where values outside of the mean $\pm$ 1.5 $\cdot$ iqr , with  iqr being the interquartile range, are threaded as outliers (BM IQR).

\begin{figure}[h]
\centering
\includegraphics[width=.45\textwidth]{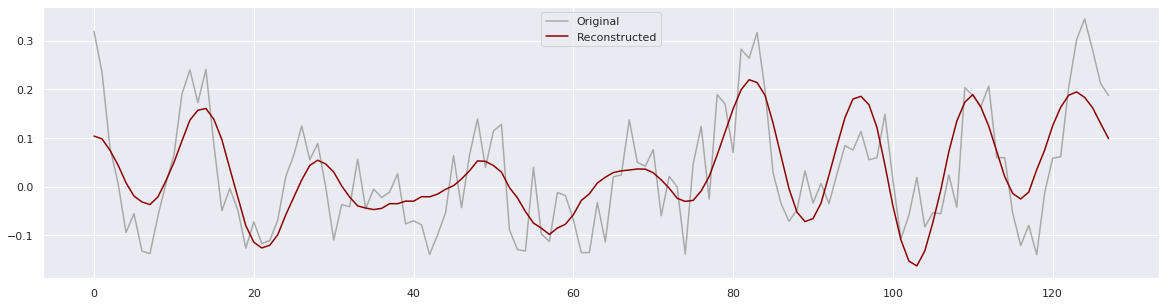}
\caption{Example of autoencoder input (measured signal) and output (reconstructed signal) from healthy data. The reconstruction error is derived input and output difference and used for anomaly detection.}
\label{fig:dataset}
\end{figure}

\section{Preliminary Results}\label{results}

We present the results of our initial experiments for all combinations of model and features in Table \ref{table_results2} and our benchmarks in Table \ref{table_results}. In terms of F1-score, which is a trade off between precision and recall, our models beat the PCA benchmark in 15 out of 24 experiments. We achieve our best result with an LSTM network and the Euclidean norm of the 3-D vibration signal (vib. 1D) resulting in a F1-score of 0.64, a precision of 0.68 and a recall of 0.6. The PCA benchmark performed best with the 3-D vibration signal in both raw and Fourier-transformed form. Thus, our best model outperformed the benchmark by 10 \%. However, the simple benchmark based on the interquartile range outperforms our autoencoders and the PCA benchmark by 1 \% with a F1-Score of 0.63.

\begin{figure}
\centering
\includegraphics[width=.45\textwidth]{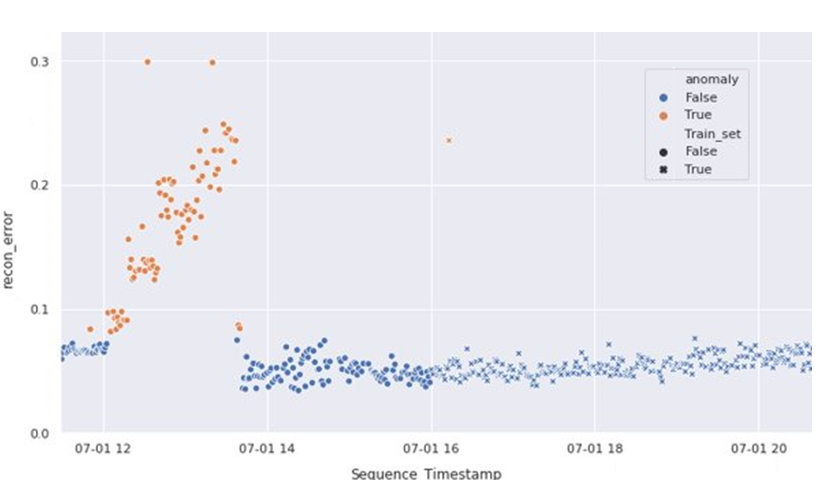}
\caption{The anomaly score is visualized for a sequence of measurements. A score above the threshold indicates an anomaly (orange). The shape of the points means whether we used it for training or evaluation of the autoencoder.}
\label{fig:dataset}
\end{figure}

\begin{table}[h]
    \centering%
    \caption{Results of benchmarks for each feature }%
    \begin{tabular}{lllll}
\hline
\multicolumn{5}{c}{Benchmark IQR} \\ \hline
Features & Acc. & F1 & P & R \\ \hline
Vibrations 1D & 0.48 & \textbf{0.63} & 0.47 & 0.94 \\
Audio & 0.56 & 0.31 & 0.55 & 0.21 \\
Vibrations 3D & 0.52 & \textbf{0.63} & 0.49 & 0.89 \\
Vibrations 1D \& Audio & 0.47 & 0.62 & 0.46 & 0.94 \\
FFT Vibrations 1D & 0.48 & \textbf{0.63} & 0.47 & 0.94 \\
FFT Audio & 0.56 & 0.31 & 0.55 & 0.21 \\
FFT Vibrations 3D & 0.52 & \textbf{0.63} & 0.49 & 0.89 \\
FFT Vibrations 1D \& Audio & 0.47 & 0.62 & 0.46 & 0.94 \\ \hline
\multicolumn{5}{c}{Benchmark PCA} \\ \hline
Features & Acc. & F1 & P & R \\ \hline
Vibrations 1D & 0.51 & 0.48 & 0.47 & 0.48 \\
Audio & 0.48 & \textbf{0.54} & 0.45 & 0.66 \\
Vibrations 3D & 0.50 & 0.45 & 0.46 & 0.44 \\
Vibrations 1D \& Audio & 0.48 & \textbf{0.54} & 0.45 & 0.66 \\
FFT Vibrations 1D & 0.51 & 0.48 & 0.47 & 0.48 \\
FFT Audio & 0.48 & \textbf{0.54} & 0.45 & 0.66 \\
FFT Vibrations 3D & 0.50 & 0.45 & 0.46 & 0.44 \\
FFT Vibrations 1D \& Audio & 0.48 & \textbf{0.54} & 0.45 & 0.66 \\ \hline
 &  &  &  & 
\end{tabular}
    \label{table_results}%
\end{table}

\begin{table} [h]
    \centering%
    \caption{Results of anomaly prediction for each combination of feature set and model.}%
    \begin{tabular}{lllll}
\hline
\multicolumn{5}{c}{convolutional neural network (CNN)} \\ \hline
Features & Acc. & F1 & P & R \\ \hline
Vibrations 1D & 0.47 & 0.47 & 0.44 & 0.50 \\
Audio & 0.46 & 0.53 & 0.44 & 0.67 \\
Vibrations 3D & 0.54 & \textbf{0.62} & 0.50 & 0.8 \\
Vibrations 1D \& Audio & 0.46 & 0.53 & 0.44 & 0.67 \\
FFT Vibrations 1D & 0.54 & 0.00 & 0.00 & 0.00 \\
FFT Audio & 0.53 & 0.46 & 0.49 & 0.44 \\
FFT Vibrations 3D & 0.54 & 0.04 & 0.57 & 0.02 \\
FFT Vibrations 1D \& Audio & 0.56 & 0.08 & 1.00 & 0.04 \\ \hline
\multicolumn{5}{c}{Fully connected neural network (DNN)} \\ \hline
Features & Acc. & F1 & P & R \\ \hline
Vibrations 1D & 0.49 & 0.53 & 0.46 & 0.61 \\
Audio & 0.46 & 0.54 & 0.45 & 0.68 \\
Vibrations 3D & 0.54 & \textbf{0.62} & 0.50 & 0.81 \\
Vibrations 1D \& Audio & 0.47 & 0.54 & 0.45 & 0.68 \\
FFT Vibrations 1D & 0.59 & 0.47 & 0.59 & 0.39 \\
FFT Audio & 0.50 & 0.45 & 0.46 & 0.44 \\
FFT Vibrations 3D & 0.49 & 0.43 & 0.44 & 0.41 \\
FFT Vibrations 1D \& Audio & 0.50 & 0.45 & 0.45 & 0.44 \\ \hline
\multicolumn{5}{c}{Long short-term memory neural network (LSTM)} \\ \hline
Features & Acc. & F1 & P & R \\ \hline
Vibrations 1D & 0.47 & 0.48 & 0.44 & 0.53 \\
Audio & 0.46 & 0.53 & 0.44 & 0.67 \\
Vibrations 3D & 0.55 & \textbf{0.62} & 0.51 & 0.79 \\
Vibrations 1D \& Audio & 0.46 & 0.53 & 0.44 & 0.67 \\
FFT Vibrations 1D & 0.53 & 0.02 & 0.40 & 0.01 \\
FFT Audio & 0.52 & 0.48 & 0.48 & 0.48 \\
FFT Vibrations 3D & 0.60 & 0.37 & 0.71 & 0.25 \\
FFT Vibrations 1D \& Audio & 0.63 & 0.42 & 0.74 & 0.30 \\ \hline
\end{tabular}
    \label{table_results2}%
\end{table}

\section{Discussion}\label{discuss}

Our aim was to define initial requirements for minimal-configuration anomaly detection for IIoT sensors. Based on the requirements, we focused on unsupervised machine learning and did not perform any equipment specific feature engineering. We  created  a  hand-crafted  benchmark  data and made it publicity available. We experimented with three with three different neural network architectures for anomaly detection in the tube system of a peristaltic pump. Our preliminary results show an important step towards minimal-configuration anomaly detection for IIoT sensors. With all three networks we were able to outperform a benchmark based on the reconstruction error of a principal component analysis. However, it remains unclear weather the the respective sensor combination can be applied to a broad number of assets. Further, the impact of sensor position and the transferability towards identical assets remain unclear. Therefore, in the future, we will perform experiments on a data set that was recorded using different sensor positions to investigate the transferability of our models.

\bibliography{conference_idsc}{}

\begin{thebibliography}{10}
\providecommand{\url}[1]{#1}
\csname url@samestyle\endcsname
\providecommand{\newblock}{\relax}
\providecommand{\bibinfo}[2]{#2}
\providecommand{\BIBentrySTDinterwordspacing}{\spaceskip=0pt\relax}
\providecommand{\BIBentryALTinterwordstretchfactor}{4}
\providecommand{\BIBentryALTinterwordspacing}{\spaceskip=\fontdimen2\font plus
\BIBentryALTinterwordstretchfactor\fontdimen3\font minus
  \fontdimen4\font\relax}
\providecommand{\BIBforeignlanguage}[2]{{%
\expandafter\ifx\csname l@#1\endcsname\relax
\typeout{** WARNING: IEEEtran.bst: No hyphenation pattern has been}%
\typeout{** loaded for the language `#1'. Using the pattern for}%
\typeout{** the default language instead.}%
\else
\language=\csname l@#1\endcsname
\fi
#2}}
\providecommand{\BIBdecl}{\relax}
\BIBdecl

\bibitem{Tolani2006}
D.~Tolani, M.~Yasar, A.~Ray, and V.~Yang, ``{Anomaly detection in aircraft gas
  turbine engines},'' \emph{Journal of Aerospace Computing, Information and
  Communication}, vol.~3, no.~2, pp. 44--51, feb 2006.

\bibitem{Zaher2009}
A.~Zaher, S.~McArthur, D.~Infield, and Y.~Patel, ``Online wind turbine fault
  detection through automated scada data analysis,'' \emph{Wind Energy: An
  International Journal for Progress and Applications in Wind Power Conversion
  Technology}, vol.~12, no.~6, pp. 574--593, 2009.

\bibitem{Tong2019}
\BIBentryALTinterwordspacing
X.~Tong, H.~Yang, L.~Wang, and Y.~Miao, ``{The Development and Field Evaluation
  of an IoT System of Low-Power Vibration for Bridge Health Monitoring},''
  \emph{Sensors}, vol.~19, no.~5, p. 1222, mar 2019. [Online]. Available:
  \url{https://www.mdpi.com/1424-8220/19/5/1222}
\BIBentrySTDinterwordspacing

\bibitem{Zhao2017}
\BIBentryALTinterwordspacing
X.~Zhao, G.~Wei, X.~Li, Y.~Qin, D.~Xu, W.~Tang, H.~Yin, X.~Wei, and L.~Jia,
  ``{Self-powered triboelectric nano vibration accelerometer based wireless
  sensor system for railway state health monitoring},'' \emph{Nano Energy},
  vol.~34, pp. 549--555, apr 2017. [Online]. Available:
  \url{https://linkinghub.elsevier.com/retrieve/pii/S2211285517301143}
\BIBentrySTDinterwordspacing

\bibitem{Bierweiler2019}
\BIBentryALTinterwordspacing
T.~Bierweiler, H.~Grieb, S.~von Dosky, and M.~Hartl, ``{Smart Sensing
  Environment – Use Cases and System for Plant Specific Monitoring and
  Optimization},'' pp. 155--158, 2019. [Online]. Available:
  \url{https://elibrary.vdi-verlag.de/index.php?doi=10.51202/9783181023518-155}
\BIBentrySTDinterwordspacing

\bibitem{Su2019}
\BIBentryALTinterwordspacing
S.~Su, Y.~Sun, X.~Gao, J.~Qiu, and Z.~Tian, ``A correlation-change based
  feature selection method for iot equipment anomaly detection,'' \emph{Applied
  Sciences}, vol.~9, no.~3, 2019. [Online]. Available:
  \url{https://www.mdpi.com/2076-3417/9/3/437}
\BIBentrySTDinterwordspacing

\bibitem{Wang2020}
\BIBentryALTinterwordspacing
J.~Wang, Y.~Tang, S.~He, C.~Zhao, P.~K. Sharma, O.~Alfarraj, and A.~Tolba,
  ``Logevent2vec: Logevent-to-vector based anomaly detection for large-scale
  logs in internet of things,'' \emph{Sensors}, vol.~20, no.~9, 2020. [Online].
  Available: \url{https://www.mdpi.com/1424-8220/20/9/2451}
\BIBentrySTDinterwordspacing

\bibitem{kato2001integrating}
Y.~Kato, T.~Yairi, and K.~Hori, ``Integrating data mining techniques and design
  information management for failure prevention,'' in \emph{Annual Conference
  of the Japanese Society for Artificial Intelligence}.\hskip 1em plus 0.5em
  minus 0.4em\relax Springer, 2001, pp. 475--480.

\bibitem{Camacho2016}
J.~Camacho, A.~P{\'{e}}rez-Villegas, P.~Garci{\'{a}}-Teodoro, and
  G.~MacI{\'{a}}-Fern{\'{a}}ndez, ``{PCA-based multivariate statistical network
  monitoring for anomaly detection},'' \emph{Computers and Security}, vol.~59,
  pp. 118--137, jun 2016.

\bibitem{Sakurada2014}
M.~Sakurada and T.~Yairi, ``Anomaly detection using autoencoders with nonlinear
  dimensionality reduction,'' in \emph{Proceedings of the MLSDA 2014 2nd
  Workshop on Machine Learning for Sensory Data Analysis}, 2014, pp. 4--11.

\bibitem{zong2018deep}
B.~Zong, Q.~Song, M.~R. Min, W.~Cheng, C.~Lumezanu, D.~Cho, and H.~Chen, ``Deep
  autoencoding gaussian mixture model for unsupervised anomaly detection,''
  2018.

\bibitem{MALEKI2021}
\BIBentryALTinterwordspacing
S.~Maleki, S.~Maleki, and N.~R. Jennings, ``Unsupervised anomaly detection with
  lstm autoencoders using statistical data-filtering,'' \emph{Applied Soft
  Computing}, vol. 108, p. 107443, 2021. [Online]. Available:
  \url{https://www.sciencedirect.com/science/article/pii/S1568494621003665}
\BIBentrySTDinterwordspacing

\end{thebibliography}
\bibliographystyle{IEEEtran}

\end{document}